# General Method for Prime-point Cyclic Convolution over the Real Field

Qi Cai, Tsung-Ching Lin, *Senior Member, IEEE*, Yuanxin Wu, *Senior Member, IEEE*,
Wenxian Yu, *Senior Member, IEEE* and Trieu-Kien Truong, *Life Fellow, IEEE*

*Abstract*—A general and fast method is conceived for computing the cyclic convolution of *n* points, where *n* is a prime number. This method fully exploits the internal structure of the cyclic matrix, and hence leads to significant reduction of the multiplication complexity in terms of CPU time by 50%, as compared with Winograd's algorithm. In this paper, we only consider the real and complex fields due to their most important applications, but in general, the idea behind this method can be extended to any finite field of interest. Clearly, it is well-known that the discrete Fourier transform (DFT) can be expressed in terms of cyclic convolution, so it can be utilized to compute the DFT when the block length is a prime.

## I. Introduction

A cyclic convolution has been widely used for a wide variety of applications in digital signal processing [1-5], image processing [6], error-correcting codes [7, 8], and synthetic aperture radar (SAR) digital processor [9, 10]. One of key applications is that cyclic convolutions over finite fields [11-13] can be employed to correct errors and erasures of Reed-Solomon (RS) codes for digital storage and deep space communication systems. Winograd's algorithm [1] is a universal standard for short cyclic convolution when $p \leq 7$ (*p* specifically denotes the prime block length of points). Then a long cyclic convolution can be obtained by piecing together short convolutions to reduce multiplicative complexity [1]. Now, let us consider the block length $n = n_1 \times n_2$ such that $n_1$ and $n_2$ are relatively prime. The short $n_1$-point and $n_2$-point cyclic convolutions can be combined to obtain a long *n*-point cyclic convolution algorithm by using the well-known Chinese remainder theorem for polynomial. However, it is very difficult or even impossible to calculate the *p*-point cyclic convolution when $p > 7$, e.g., see Table 5.4 in [1].

It is well-known that if the number of points *n* is a prime (denoted as *p*), using the Rader prime algorithm [2], the discrete Fourier transform (DFT) can be expressed in terms of cyclic convolutions. They are computed by a use of the Winograd short cyclic convolution algorithm [3]. As mentioned earlier in [1], it can also be viewed as a method of factoring a cyclic matrix. For example, when a prime number of points *p*=859, the DFT can be expressed in terms of a (*p*-1=858)-point cyclic convolution. Since $858 = 2 \times 3 \times 11 \times 13$, where 2, 3, 11, 13 are all prime numbers, this 858-point cyclic convolution could be obtained by using cyclic convolutions of 2, 3, 11, and 13 points [3] only if the *p*-point cyclic convolution problem for $p > 7$ was solved.

To overcome the large prime problem, in this paper, a general and fast method is derived to calculate the *p*-point cyclic convolution by fully exploiting the internal structure of the cyclic matrix, which is inspired by Winograd's method. Obviously, the direct computation involves $n^2$ multiplications and $n(n-1)$ additions. However, the number of multiplications and additions needed to compute the general method are reduced to $n(n-1)/2+1$ and $3n(n-1)/2+1$, respectively.

The remainder of this paper is organized as follows: Some backgrounds of constructing cyclic convolutions are presented in Section II. In Section III, Winograd's short cyclic convolution is briefly described. Section IV explores a fast method to compute the *p*-point cyclic convolution. The computational complexity of the new method is analyzed, and then compared with other existing state-of-the-art algorithms. Simulation results of the proposed approach are presented in Section V. Finally, this paper concludes with a brief summary in Section VI.

## II. Definitions

In this section, we describe some definitions of a cyclic convolution, which will be needed in developing both the Winograd short cyclic convolution algorithm and the new algorithm proposed in this paper.

First, let three vectors of *n* points be $z = (z_0, \ldots, z_{n-1})^T$, $c = (c_0, \ldots, c_{n-1})^T$, and $b = (b_0, \ldots, b_{n-1})^T$. Then the cyclic convolution of two sequences $b$ and $y$ can be written as

This work was supported in part by National Natural Science Foundation of China (61673263) and National Key R&D Program of China (2018YFB1305103).

Q. Cai, Y. Wu, W. Yu and T. K. Truong are with Shanghai Key Laboratory of Navigation and Location-based Services, School of Electronic Information and Electrical Engineering, Shanghai Jiao Tong University, Shanghai, China, 200240, (e-mail: qicaiCN@gmail.com; yuanx_wu@hotmail.com; wxyu@sjtu.edu.cn, truong@isu.edu.tw).

T.C. Lin is with Shanghai Key Laboratory of Navigation and Location-based Services, School of Electronic Information and Electrical Engineering, Shanghai Jiao Tong University, Shanghai, China, 200240; and also with Department of Information Engineering, I-Shou University, Kaohsiung City 84001, Taiwan, R.O.C. (email: joe@isu.edu.tw).

$$c_p = \sum_{l=0}^{n-1} b_l z_{(p-l)_n} \text{ for } 0 \leq p \leq n-1 , \tag{1}$$

where $(p-l)_n$ denotes the residue of $p-l$ module $n$.

Next, Eq. (1) can be written in the form of a matrix as follows:

$$\begin{pmatrix} c_0 \\ c_1 \\ \vdots \\ c_{n-1} \end{pmatrix} = \begin{pmatrix} b_0 & b_1 & \cdots & b_{n-1} \\ b_1 & b_2 & \cdots & b_0 \\ \vdots & \vdots & \ddots & \vdots \\ b_{n-1} & b_0 & \cdots & b_{n-2} \end{pmatrix} \begin{pmatrix} z_0 \\ z_{n-1} \\ \vdots \\ z_1 \end{pmatrix} \tag{2}$$

or

$$\boldsymbol{c} = B^c \boldsymbol{y} , \tag{3}$$

where $\boldsymbol{y} = (y_0,\ldots,y_{n-1})^T = (z_0, z_{n-1},\ldots,z_1)^T$ and $B^c$ denotes a cyclic matrix whose components are known a priori. If $m(x) = x^n - 1$ is a fixed polynomial of degree $n$, it can be readily verified that the foregoing matrix can also be expressed more explicitly in terms of the coefficients of the polynomial [1], given by

$$c(x) \equiv b(x) z(x) \bmod x^n - 1 , \tag{4}$$

where $c(x) = c_0 x^0 + \cdots + c_{n-1} x^{n-1}$, $b(x) = b_0 x^0 + \cdots + b_{n-1} x^{n-1}$, and $z(x) = z_0 x^0 + z_1 x^1 \cdots + z_{n-1} x^{n-1}$ are three polynomials of degree $n$-1 whose coefficients are in the real or complex field.

In addition, a linear convolution of two sequences $b_l$ and $z_l$ for $0 \leq l \leq n-1$ is defined as

$$c_p = \sum_{l=0}^{n-1} b_l z_{(p-l)} \text{ for } 0 \leq p \leq n-1 , \tag{5}$$

where $c_p$ is a finite duration sequence of $n$ samples.

It is well-known that if one appends zero-valued samples to both $b_l$ and $z_l$ such that they are two $2n$-point sequences, clearly, a cyclic convolution of these two $2n$-point sequences can be obtained by using a $2n$-point DFT. Thereby, we obtain the linear convolution $c_p$ as shown in (5) if we only take the first $n$-point sequence of the cyclic convolution result, see [1].

### III. WINOGRAD SHORT CYCLIC CONVOLUTION ALGORITHM

The following is a brief review of Winograd's algorithm. For more detail, the interested reader is referred to [1, 3].

In order to illustrate the Winograd short cyclic convolution algorithm, the following theorem will be needed.

*Chinese Remainder Theorem [7]*. Let $m(x)$ be factored into pairwise coprime polynomials $m_k(x)$ for $0 \leq k \leq K-1$ in the real field; that is,

$$m(x) = m_0(x) m_1(x) \cdots m_{K-1}(x). \tag{6}$$

With these factorizations, evaluating the residue polynomials yields

$$\begin{aligned} c_k(x) &\equiv c(x) \bmod m_k(x) \\ &\equiv b(x) z(x) \bmod m_k(x). \end{aligned} \tag{7}$$

Then the system of congruences given in (7) has a unique solution $c(x) \bmod m(x)$, given by

$$c(x) \equiv \sum_{k=0}^{K-1} c_k(x) M_k(x) M_k^{-1}(x) \bmod m(x) , \tag{8}$$

where $M_k(x) = m(x)/m_k(x)$ and $M_k^{-1}(x)$ uniquely satisfies the congruence

$$M_k(x) M_k^{-1}(x) \equiv 1 \bmod m_k(x) . \tag{9}$$

**Proof.**

Since $GCD(m_k(x), M_k(x)) = 1$, by the Euclidean algorithm [7], there is a unique solution $M_k^{-1}(x)$ modulo $m_k(x)$ of (9). If $c(x)$ is given in (8), then compute the residues

$$\begin{aligned} c(x) &\equiv c_k(x) M_k(x) M_k^{-1}(x) \\ &\equiv c_k(x) \cdot 1 \\ &\equiv c_k(x) \bmod m_k(x) \text{ for every } k. \end{aligned} \tag{10}$$

Suppose $d(x)$ is another solution satisfying (7). Then there is

$$\begin{aligned} d(x) - c(x) &\equiv (b(x) z(x) - b(x) z(x)) \bmod m_k(x) \\ &\equiv 0 \bmod m_k(x). \end{aligned} \tag{11}$$

This implies that $d(x) \equiv c(x) \bmod m(x)$. Since $m_k(x)$'s are coprime polynomials. The solution of (8) is thus unique.
Q.E.D.

It is easy to see that $M_k(x) M_k^{-1}(x)$ in (8) is a polynomial with rational coefficients. Hence, the computation of (8) involves no multiplications at all.

It is of interest to note that if $n$ is a composite number, then the polynomial $x^n - 1$ may factor into more prime polynomials. Finally, the Chinese remainder theorem will make a natural use of these residue polynomials so as to dramatically reduce the number of multiplications needed in computing the *n*-point cyclic convolution. Unfortunately, if $n$ is a prime, i.e., $n=p$, we only factor $x^p - 1$ into two prime polynomials over the real field; that is, $x^p - 1 = m_0(x) m_1(x)$, where $m_0(x) = x - 1$ and $m_1(x) = x^{p-1} + x^{p-2} + \cdots + x + 1$. To compute the polynomial product modulo $x^p - 1$, namely $c(x) \equiv b(x) z(x) \bmod (x^p - 1)$, one first needs to compute the following two residue polynomials:

$$c_0(x) \equiv c(x) \bmod (x-1) \tag{12}$$

and

$$c_1(x) \equiv c(x) \bmod (x^{p-1} + x^{p-2} + \cdots + x + 1) . \tag{13}$$

The computation of (12) is obvious and requires only one multiplication. However, it is shown in [1] that computing (13)

involves $\left[\text{degree } m_1(x)\right]^2 = (p-1)^2$ multiplications, and hence this approach results in high multiplicative complexity when $p$ is large. To make the computational complexity even less for (13), a good algorithm needed to compute the cyclic convolution of any prime number of points will be constructed in the next section.

Specifically, the following theorem is very important for finding the minimum bound of multiplications of the cyclic convolutions of two sets of $p$ points.

**Theorem 1.** Let $m(x) = x^p - 1$. The minimum number of multiplications needed in computing the polynomial product
$$c(x) \equiv b(x)z(x) \bmod (x^p - 1) \quad (14)$$
is $2(p-1)$.

**Proof.**
First, the residue polynomials are computed. In this case, the calculation of (12) requires only one multiplication. Since $x^{p-1} + x^{p-2} + \cdots + 1$ is a prime polynomial of degree $p-1$, by the use of Theorem 5.8.4 in [1], every algorithm needed to compute (13) requires at least $2(p-1)-1$ multiplications, so the total number of multiplications needed in computing (14) is $1+2(p-1)-1 = 2(p-1)$.
Q.E.D.

It is worth noting that, to our best knowledge, an algorithm has never been seen in the literature in the past decades in that it can achieve such a minimum bound for cyclic convolution computations except $p = 3, 5$ as shown in Table 5.4 in [1].

## IV. P-POINT CYCLIC CONVOLUTION ALGORITHM

In what follows, a novel algorithm with reduced multiplicative complexity is derived to compute the cyclic convolution of any prime number of points. The inspiration of our idea comes from Winograd's method.

As mentioned earlier, the cyclic convolution of two sequences $b$ and $y$ are given in (3). This equation can be rewritten by the following manner:
$$c = B^c y = \left(b \quad D^T b \quad (D^T)^2 b \quad \cdots \quad (D^T)^{n-1} b\right) y, \quad (15)$$

where $B^c$ is expressed in terms of its column vectors $(D^T)^i b$ for $i = 0, \ldots, n-1$ and $D = (e_2 \quad \cdots \quad e_n \quad e_1)$ whose column vector $e_i$ is the i-th column of an identity matrix, i.e.,
$$D = \begin{pmatrix} 0 & 0 & 0 & 0 & 1 \\ 1 & 0 & 0 & 0 & 0 \\ 0 & 1 & 0 & 0 & 0 \\ \vdots & \vdots & \ddots & \vdots & 0 \\ 0 & 0 & 0 & 1 & 0 \end{pmatrix}. \quad (16)$$

Obviously, (15) is shown to be
$$c = \begin{pmatrix} b^T y & b^T Dy & \cdots & b^T D^{n-1} y \end{pmatrix}^T. \quad (17)$$

In the sequel, let $d = (1, \ldots, 1)^T$ be a column vector of $n$ points.

**Property 1.** $D^n = I_n$, where $I_n$ is an identity matrix of size $n$.
**Proof.**
Since $D^{n-1} = (e_n \quad e_1 \quad \cdots \quad e_1)$, then there is $DD^{n-1} = (e_1 \quad e_2 \quad \cdots \quad e_n) = I_n$.
Q.E.D.

**Property 2.** $\sum_{i=0}^{n-1} D^i = dd^T$ for $i = 0, \ldots, n-1$.

**Proof.**
According to the proof of Property 1, the sum of $D^i$ for $i = 0, \ldots, n-1$ can be derived as
$$\sum_{i=0}^{n-1} D^i = \left(\sum_{i=0}^{n-1} e_i \quad \cdots \quad \sum_{i=0}^{n-1} e_i\right)$$
$$= \sum_{i=0}^{n-1} e_i (1 \quad \cdots \quad 1) = dd^T \quad (18)$$
Q.E.D.

Clearly, left multiplying (15) by a vector $s^T = (s_0 \quad \cdots \quad s_{n-1})$ yields
$$s^T c = s^T B^c y = \left(b^T D^0 s \quad \cdots \quad b^T D^{n-1} s\right) y$$
$$= b^T \left(D^0 s \quad \cdots \quad D^{n-1} s\right) y = b^T F y, \quad (19)$$
where $F = \left(D^0 s \quad \cdots \quad D^{n-1} s\right) \triangleq F(s)$ is a matrix of size $n \times n$.

In particular, to reduce the number of multiplications for $b^T F y$, two theorems need to be derived as follows:

**Theorem 2.** The number of multiplicative operations in $b^T F y$ is at least $rank(F)$.
**Proof.**
Assume that $b^T F y$ can be obtained by at least $m$ multiplicative operations. Using (19), there always exist $m$ vectors $h_i$ and $g_i$ for $i = 0, \ldots, m-1$ such that
$$b^T F y = \sum_{i=0}^{m-1} \left(b^T h_i\right)\left(g_i^T y\right) = b^T \left(\sum_{i=0}^{m-1} h_i g_i^T\right) y, \quad (20)$$
where the elements of the vector $g_i$ must be one of -1, 0, and 1.

Since $rank(A+B) \leq rank(A) + rank(B)$, one has $rank(F) = rank\left(h_1 g_1^T + \cdots + h_m g_m^T\right) \leq m$ for which $rank\left(h_i g_i^T\right) \leq 1$. Therefore, the number of multiplications $m$ is at least $rank(F)$.
Q.E.D.

**Theorem 3.** If $d^T s = 0$, $F$ is a singular matrix and the sum of its column vectors is equivalent to zero vector; that is, $s + Ds + \cdots + D^{n-1}s = 0$.
**Proof.**

Considering that $d^T s = 0$, the sum of all column vectors of the matrix $F$ is
$$s + Ds + \cdots + D^{n-1}s = dd^T s = 0 . \quad (21)$$
This implies that $F$ is a singular matrix with $rank(F) \leq n-1$.
Q.E.D.

According to Theorems 2-3, two $n \times 1$ vectors $d = (1, \cdots, 1)^T$ and $s = (-n+1, 1, \cdots, 1)^T$ lead to the simplification of multiplicative complexity in (3). It is obvious to observe that $d^T s = 0$ and $d = Dd = \cdots = D^{n-1}d$.

Meanwhile, left multiplying (3) by both $(D^i s)^T$ for $i = 0, \ldots, n$-1 and $d^T$. And applying the result of (19), we have
$$(D^i s)^T c = b^T D^i F y \quad (22)$$
and
$$d^T c = b^T dd^T y . \quad (23)$$
Let $h = (h_0 \cdots h_{n-1})^T$ whose elements are $h_i = b^T D^i F y$ ($i = 0, \ldots, n$-1). Then three important propositions are given below:

**Proposition 1.** $d^T h = 0$.
**Proof.**
Considering that $h_i = b^T D^i F y$, we have

$$d^T h = \sum_{i=0}^{n-1} b^T D^i F y = b^T \left( \sum_{i=0}^{n-1} D^i \right) F y = b^T dd^T F y = 0 , \quad (24)$$

where $d^T F = 0_{1 \times n}$ because of $d^T D^i s = 0$.
Q.E.D.

**Proposition 2.** $I_n = (dd^T - F)/n$.
**Proof.**
The matrix $F$ is of the form
$$F = \begin{pmatrix} -n+1 & 1 & \cdots & 1 \\ 1 & -n+1 & \cdots & 1 \\ \vdots & \vdots & \ddots & \vdots \\ 1 & 1 & \cdots & -n+1 \end{pmatrix}.$$
Thus, $I_n = (dd^T - F)/n$ holds obviously.
Q.E.D.

**Proposition 3.** The output vector can be calculated by means of $c = (b^T dd^T y - h)/n$.
**Proof.**
According to Proposition 2, for the i-th element of $c$, we know that
$$\begin{aligned} b^T D^i y &= \frac{1}{n} b^T D^i (dd^T - F) y \\ &= \frac{1}{n} b^T dd^T y - \frac{1}{n} h_i . \end{aligned} \quad (25)$$
The output vector $c$ immediately follows from (17).
Q.E.D

Proposition 3 shows that the calculation of $c$ needs $b^T dd^T y$ and $h$. Furthermore, let $h' = (h_0 \cdots h_{n-2})^T$, Proposition 1 indicates that $h$ can be easily obtained by $h'$.

In conclusion, it is necessary to calculate both $b^T dd^T y$ and $h'$ for the output vector $c$ where $b^T dd^T y$ is known and $h'$ can be expressed as

$$h' = \begin{pmatrix} h_0 \\ \vdots \\ h_{n-2} \end{pmatrix} = \begin{pmatrix} b^T F y \\ \vdots \\ b^T D^{n-2} F y \end{pmatrix} = \begin{pmatrix} b^T D^0 s & \cdots & b^T D^{n-1} s \\ \vdots & \ddots & \vdots \\ b^T D^{n-2} s & \cdots & b^T D^{n-3} s \end{pmatrix} y . \quad (26)$$

In accordance with Theorem 3, we have

$$h' = \begin{pmatrix} -b^T \sum_{i=1}^{n-1} D^i s & b^T Ds & \cdots & b^T D^{n-2} s & b^T D^{n-1} s \\ b^T Ds & -b^T \sum_{\substack{i=0 \\ i \neq 1}}^{n-1} D^i s & \cdots & b^T D^{n-1} s & b^T s \\ \vdots & \vdots & \ddots & \vdots & \vdots \\ b^T D^{n-2} s & b^T D^{n-1} s & \cdots & -b^T \sum_{\substack{i=0 \\ i \neq n-2}}^{n-1} D^i s & b^T D^{n-3} s \end{pmatrix} y. \quad (27)$$

Here, $h'$ can be further reduced as
$$h_i = \sum_{j=0}^{n-1} g_{i,j} , \quad (28)$$
where $g_{i,j} = (b^T D^{i+j} s)(y_j - y_i)$ for $i = 0, \ldots, n$-2 and $j = 0, \ldots, n$-1. For each $g_{i,j}$, it requires only one multiplication and one addition, and then $n$ multiplications for each $h_i$ as given in (28) is needed. In fact, since $g_{i,i} = 0$, the number of multiplications for $h_i$ is reduced from $n$ to $n$-1, which is equal to the rank of the matrix $F$ and is the minimum multiplication number for each $h_i$ according to Theorem 2. As a result, the computation of $h'$ given in (26) requires $(n-1)^2$ multiplications.

TABLE I.
PERFORMANCE OF THE P-POINT CYCLIC CONVOLUTION ALGORITHMS

| Block Length $p$ | Best Algorithms [1] | | General Algorithm | | Direct Algorithm | |
|---|---|---|---|---|---|---|
| | M | A | M $n(n-1)/2+1$ | A $3n(n-1)/2+1$ | M $n^2$ | A $n(n-1)$ |
| 3 | 4 | 11 | 4 | 10 | 9 | 6 |
| 5 | 8 | 62 | 11 | 31 | 25 | 20 |
| 7 | 16 | 70 | 22 | 64 | 49 | 42 |
| 11 | | | 56 | 166 | 121 | 110 |
| 13 | | | 79 | 235 | 169 | 156 |
| 17 | | | 137 | 409 | 189 | 172 |

| 19 | 172 | 514 | 361 | 342 |
| 23 | 254 | 760 | 529 | 506 |

Fortunately, considering the relationship between $h_i$ and $h_j$, there always exists an important relationship $g_{i,j} = -g_{j,i}$ in such a way that one only need to compute the part of $g_{i,j}$ with $i < j$ to get $\boldsymbol{h'}$. Thus, the number of multiplications and additions needed to obtain $\boldsymbol{h'}$ are $(n-1)n/2$ and $(n-1)(3n-4)/2$, respectively. Whereas, computing $\boldsymbol{h}$ involves $(n-1)n/2$ multiplications and $(n-1)(3n-2)/2$ additions. Additionally, the calculation of $\boldsymbol{b}^T\boldsymbol{dd}^T\boldsymbol{y}$ uses only one multiplication and $n$-1 addition.

In conclusion, according to Proposition 3, $(n-1)n/2+1$ multiplications and $3n(n-1)/2+1$ additions are needed to obtain the output vector $\boldsymbol{c}$. For different prime points of cyclic convolutions, the number of multiplications and additions needed in the general algorithm is subtly listed in Table I and compared to the direct computation and other state-of-the-art approaches. In this table, for simplicity, M and A denote the number of multiplications and additions, respectively. It should be noted that no efficient algorithms are found to compute the cyclic convolutions of a prime number of points greater than seven except our algorithm.

The proposed general algorithm for computing the $p$-point cyclic convolutions is composed of the following four steps:

**Step 1.** Pre-calculate $v_i \triangleq \boldsymbol{b}^T D^i \boldsymbol{s}/n, 0 \leq i \leq n\text{-}1$.

**Step 2.** Calculate $l_{i,j} \triangleq (y_j - y_i)$, and then calculate $g_{i,j} = v_{j+i}l_{i,j}$ when $i+j \leq n\text{-}1$; otherwise, $g_{i,j} = v_{j+i-n}l_{i,j}$ for $0 \leq i \leq n-2$ and $i < j \leq n-1$.

**Step 3.** Calculate $q = \frac{1}{n}\boldsymbol{b}^T\boldsymbol{dd}^T\boldsymbol{y}$ and

$$h_i = \begin{cases} \sum_{j=i+1}^{n-1} g_{i,j} - \sum_{j=0}^{i-1} g_{j,i} & \text{for } 0 \leq i \leq n-2 \text{ and } 0 \leq j \leq n-1 \\ -\sum_{i=0}^{n-2} h_i & \text{for } i = n-1. \end{cases}$$

**Step 4.** Obtain the output vector $\boldsymbol{c}$ by $c_i = q - h_i$ for $1 \leq i \leq n$.

The algorithm described above is illustrated by an example bellow:

For the case of $n = 3$, $\boldsymbol{d} = (1,1,1)^T$ and $\boldsymbol{s} = (-2,1,1)^T$. Eq.(2) becomes

$$\begin{pmatrix} c_0 \\ c_1 \\ c_2 \end{pmatrix} = \begin{pmatrix} b_0 & b_1 & b_2 \\ b_1 & b_2 & b_0 \\ b_2 & b_0 & b_1 \end{pmatrix} \begin{pmatrix} y_0 \\ y_1 \\ y_2 \end{pmatrix}, \quad (29)$$

where $y_0 = z_0, y_1 = z_2,$ and $y_2 = z_1$.
Then compute

$$F = \begin{pmatrix} D^0\boldsymbol{s} & D^1\boldsymbol{s} & D^2\boldsymbol{s} \end{pmatrix} = \begin{pmatrix} -2 & 1 & 1 \\ 1 & -2 & 1 \\ 1 & 1 & -2 \end{pmatrix}. \quad (30)$$

With the aid of Proposition 3, the output vector $\boldsymbol{c}$ should have the form

$$c_i = \frac{1}{3}\boldsymbol{b}^T\boldsymbol{dd}^T\boldsymbol{y} - h_i, \quad (31)$$

where $h_i = \frac{1}{3}\boldsymbol{b}^T D^i F\boldsymbol{y}$ for $i = 1, 2, 3$. It means that the vector $\boldsymbol{c}$ can be obtained by means of the known $\frac{1}{3}\boldsymbol{b}^T\boldsymbol{dd}^T\boldsymbol{y}$ and $\boldsymbol{h} = (h_0 \ h_1 \ h_2)^T$. The number of multiplications and additions in (31) are, respectively, 1 and 4. The next point that we concern is how to efficiently calculate $\boldsymbol{h}$.

TABLE II
ELEMENTS OF CALCULATING $\boldsymbol{h'}$

| Elements of $v$ | Elements of $l$ | Elements of $g$ |
|---|---|---|
| $v_0 = \boldsymbol{b}^T D^0 \boldsymbol{s}/3$ | $l_{0,1} = y_1 - y_0$ | $g_{0,1} = v_1 l_{0,1}$ |
| $v_1 = \boldsymbol{b}^T D^1 \boldsymbol{s}/3$ | $l_{0,2} = y_2 - y_0$ | $g_{0,2} = v_2 l_{0,2}$ |
| $v_2 = \boldsymbol{b}^T D^2 \boldsymbol{s}/3$ | $l_{1,2} = y_2 - y_1$ | $g_{1,2} = v_0 l_{1,2}$ |
| | | $g_{1,0} = -g_{0,1}$ |

Firstly, one can see that $\boldsymbol{d}^T\boldsymbol{h}$ is always equal to zero so that one can only calculate $\boldsymbol{h'} = (h_0 \ h_1)^T$, thereby obtaining $h_2$ by $-\sum_{i=0}^{1} h_i$. This will lead to one more extra addition. Finally, we focus on the calculation of $\boldsymbol{h'}$. That is,

$$\boldsymbol{h'} = \begin{pmatrix} h_0 \\ h_1 \end{pmatrix} = \begin{pmatrix} \frac{1}{3}\boldsymbol{b}^T D^0 F\boldsymbol{y} \\ \frac{1}{3}\boldsymbol{b}^T D^1 F\boldsymbol{y} \end{pmatrix} = \begin{pmatrix} g_{0,1} + g_{0,2} \\ g_{1,0} + g_{1,2} \end{pmatrix}, \quad (32)$$

where the necessary elements are listed in Table II. The calculation of $\boldsymbol{h'}$ uses three multiplications (i.e., $g_{0,1}$, $g_{0,2}$, and $g_{1,2}$) and five additions.

Generally, the number of multiplications and additions needed to compute the output vector $\boldsymbol{c}$ are 1+3=4 and 4+1+5=10, respectively.

## V. CONCLUSIONS

Short $p$-point cyclic convolutions over the real and complex fields were first essentially derived by Winograd. It makes a use of the Chinese remainder theorem and enables one to derive an efficient algorithm for $n$-point cyclic convolutions. In this paper, a general and fast method is conceived to compute any $p$-point cyclic convolution over the real field. The analysis of computational complexity indicates that its speed is over 50% percent faster than that of the direct method. Finally, it is expected that this algorithm may further be improved to compute any *p*-point cyclic convolutions with fewer or at least 2(*p*-1) multiplications.

ACKNOWLEDGMENT

The authors would like to thank Dr. Jack Chang for his useful comments and suggestions of the research that led to this paper.